\documentclass[10pt,twocolumn,letterpaper]{article}

\usepackage{cvpr}
\usepackage{times}
\usepackage{epsfig}
\usepackage{graphicx}
\usepackage{amsmath}
\usepackage{nopageno}
\usepackage{amssymb}
\usepackage{booktabs}
\usepackage{algorithm}
\usepackage{algorithmic}
\usepackage{subcaption}
\usepackage{adjustbox}
\usepackage{tabu,multirow}
\usepackage{amsthm}
\usepackage{float}
\usepackage[normalem]{ulem}
\usepackage{pifont}

\newtheorem{thm}{Theorem} 

\usepackage{enumitem}
\usepackage[breaklinks=true,bookmarks=false]{hyperref}

\usepackage[all]{nowidow}
\cvprfinalcopy 

\def\cvprPaperID{5942} 

\begin{document}

\title{Enhancing Intrinsic Adversarial Robustness via Feature Pyramid Decoder\footnotemark[3]}
\author{Guanlin Li$^{1,}$\footnotemark[1]\qquad Shuya Ding$^{2,}$\footnotemark[1]\qquad Jun Luo$^2$ \qquad Chang Liu$^2$\\
\!\!\!\!\!\!{\small $^1$Shandong Provincial Key Laboratory of Computer Networks, Shandong Computer Science Center (National Supercomputer Center in Jinan)}\\
{\small \ $^2$School of Computer Science and Engineering, Nanyang Technological University}\\
{\tt\small leegl@sdas.org\qquad\{di0002ya,junluo,chang015\}@ntu.edu.sg}
}

\maketitle

\renewcommand{\thefootnote}{\fnsymbol{footnote}} 
\footnotetext[1]{These authors contributed equally to this work.} 
\footnotetext[2]{The author is also affiliated with Shandong Computer Science Center, Shandong Academy of Sciences, School of Cyber Security, Qilu University of Technology, China}
\footnotetext[3]{https://github.com/GuanlinLee/FPD-for-Adversarial-Robustness}

\begin{abstract}

%
Whereas adversarial training is employed as the main defence strategy against specific adversarial samples, it has limited generalization capability and incurs excessive time complexity. In this paper, we propose an attack-agnostic defence framework to enhance the intrinsic robustness of neural networks, without jeopardizing the ability of generalizing clean samples. Our Feature Pyramid Decoder (FPD) framework applies to all block-based convolutional neural networks (CNNs). It implants denoising and image restoration modules into a targeted CNN, and it also constraints the Lipschitz constant of the classification layer. Moreover, we propose a two-phase strategy to train the FPD-enhanced CNN, utilizing $\epsilon$-neighbourhood noisy images with multi-task and self-supervised learning. Evaluated against a variety of white-box and black-box attacks, we demonstrate that FPD-enhanced CNNs gain sufficient robustness against general adversarial samples on MNIST, SVHN and CALTECH. In addition, if we further conduct adversarial training, the FPD-enhanced CNNs perform better than their non-enhanced versions.

\end{abstract}
\section{Introduction}

The ever-growing ability of deep learning has found numerous applications mainly in image classification, object detection, and natural language processing~\cite{he_deep_2016,lin_feature_2017,vaswani_attention_2017}. While deep learning has brought great convenience to our lives, its weakness is also catching researchers' attention. Recently, researchers have started to pay more attention to investigating the weakness of neural networks, especially in its application to image classification. 
Since the seminal work by~\cite{szegedy_intriguing_2013,nguyen_deep_2015}, many follow-up works have demonstrated a great variety of methods in generating \textit{adversarial samples}: though easily distinguishable by human eyes, they are often misclassified by neural networks. More specifically, most convolutional layers are very sensitive to perturbations brought by adversarial samples (\eg,~\cite{che_adversarial_2019,xu_interpreting_2019}), resulting in misclassifications. These so-called \textit{adversarial attacks} may adopt either white-box or black-box approaches, depending on the knowledge of the target network, and they mostly use gradient-based methods~\cite{goodfellow_explaining_2014,madry_towards_2017,tramer_ensemble_2018} or score-based methods~\cite{carlini_towards_2017} to generate adversarial samples. 

To thwart these attacks, many defence methods have been proposed. Most of them use \textit{adversarial training} to increase the network robustness, \eg,~\cite{athalye_obfuscated_2018,kannan_adversarial_2018}. However, as training often targets a specific attack, the resulting defense method can hardly be generalized, as hinted in~\cite{tramer_ensemble_2018}. In order to defend against various attacks, a large amount and variety of adversarial samples are required to retrain the classifier, leading to a high time-complexity. In the meantime, little attention has been given to the direct design of robust frameworks in an \textit{attack-agnostic} manner, except a few touches on denoising~\cite{song_pixeldefend:_2018,xie_feature_2019} and obfuscating gradients~\cite{guo_countering_2018,song_pixeldefend:_2018} that aim to directly enhance a target network in order to cope with any potential attacks.



To enhance the intrinsic robustness of neural networks, we propose an attack-agnostic defence framework, applicable to enhance all types of block-based CNNs. We aim to thwart both white-box and black-box attacks without crafting any specific adversarial attacks. Our Feature Pyramid Decoder (FPD) framework implants a target CNN with both denoising and image restoration modules to filter an input image at multiple levels; it also deploys a Lipschitz Constant Constraint at the classification layer to limit the output variation in the face of attack perturbation. In order to train an FPD-enhanced CNN, we propose a two-phase strategy; it utilizes $\epsilon$-neighbourhood noisy images to drive multi-task and self-supervised learning.
\begin{figure*}[ht]
\centering
\includegraphics[trim={3cm 4cm 3cm 5cm}, width=0.8\textwidth]{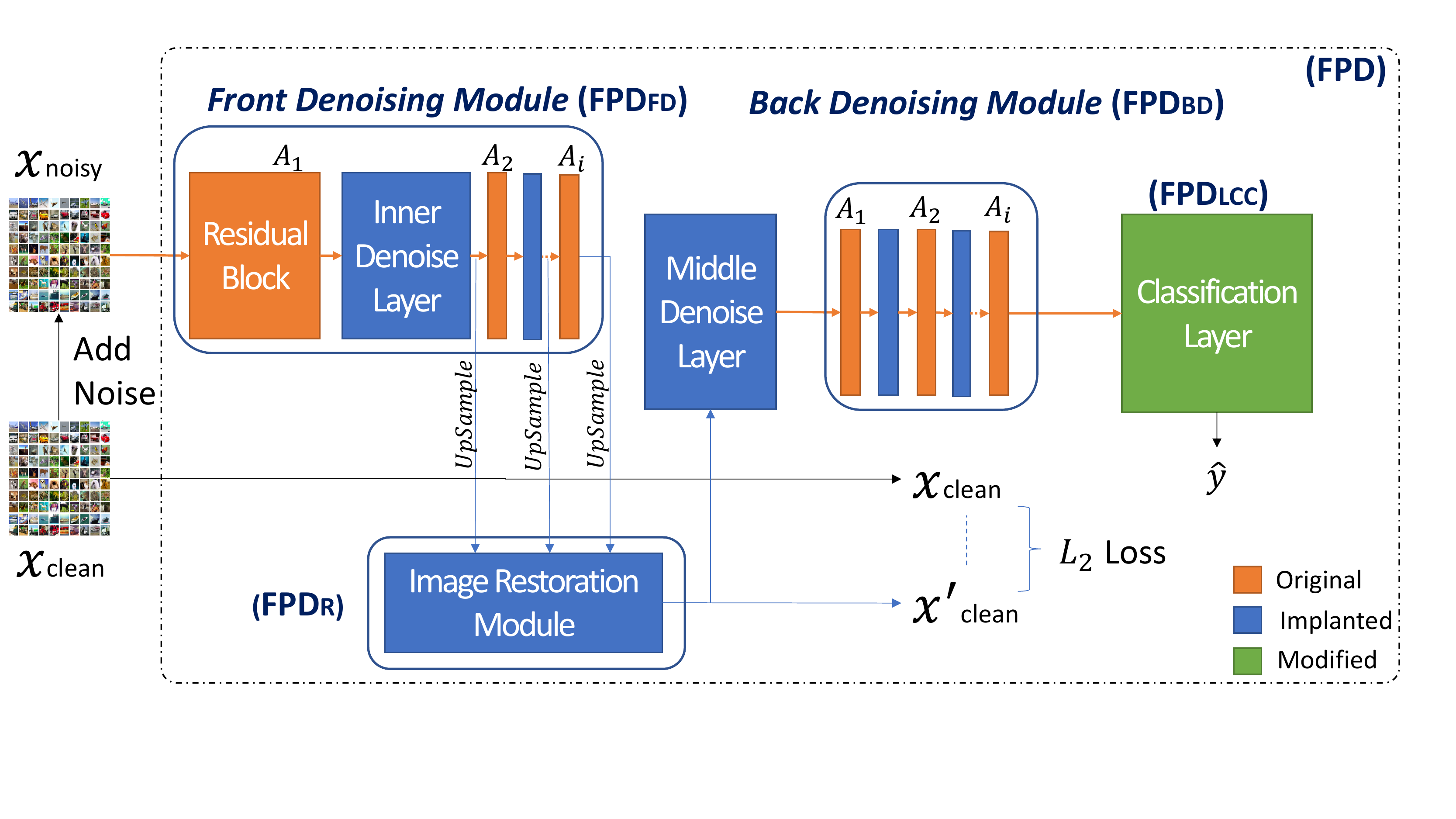}
\vspace{4ex}
\caption{
The structure of the block-based CNN, enhanced with the proposed framework named FPD: it consists of the Lipschitz constant constrained classification layer $\mathrm{FPD}_\mathrm{LCC}$; the front denoising module $\mathrm{FPD}_\mathrm{FD}$, the image restoration module $\mathrm{FPD}_\mathrm{R}$, a middle denoising layer and the back denoising module $\mathrm{FPD}_\mathrm{BD}$. $\epsilon$-neighbourhood noisy samples $x_\mathrm{nosiy}$ and original samples $x_\mathrm{clean}$ are used to train the FPD. Orange, blue and green blocks represent the original components of the CNN, the proposed components that implanted to the CNN, the modified components of the CNN, respectively.}
\label{fig:hl}
\vspace{-1ex}
\end{figure*}

As shown in Figure~\ref{fig:hl}, FPD employs a front denoising module, an image restoration module, a middle denoising layer, and a back denoising module. Both front and back denoising modules consist of the original CNN blocks interleaved with inner denoising layers, and the inner denoising layers are empirically implanted only to the shallow blocks of the CNN. Enabled by the image restoration module, the whole enhanced CNN exhibits a multi-scale pyramid structure. The multi-task learning concentrates on improving both the quality of the regenerate images $x'_\mathrm{clean}$ and the performance of final classification. Aided by the supervision target $x_\mathrm{clean}$, the enhanced CNN could be trained to denoise images and abstract the features from the denoised images. 
%
%
%
%
In summary, we make the following major contributions:
\begin{itemize}[noitemsep]
    \item Through a series of exploration experiments, we propose a novel defence framework. Our FPD framework aims to enhance the intrinsic robustness of all types of block-based CNN.
    \item We propose a two-phase strategy for strategically training the enhanced CNN, utilizing $\epsilon$-neighbourhood noisy images with both self-supervised and multi-task learning. %
    \item We validate our framework performance on both MNIST, SVHN and CALTECH datasets in defending against a variety of white-box and black-box attacks, achieving promising results. Moreover, under adversarial training, an enhanced CNN is much more robust than the non-enhanced version. 
\end{itemize}

Owing to unavoidable limitations of evaluating robustness, we release our network in \href{https://github.com/GuanlinLee/FPD-for-Adversarial-Robustness}{github\footnotemark[3]} to invite researchers to conduct extended evaluations.

\begin{figure*}[h!]
    \centering
    \begin{subfigure}[ht]{0.33\textwidth}
        \centering
        \includegraphics[width=0.5\columnwidth,height=138pt]{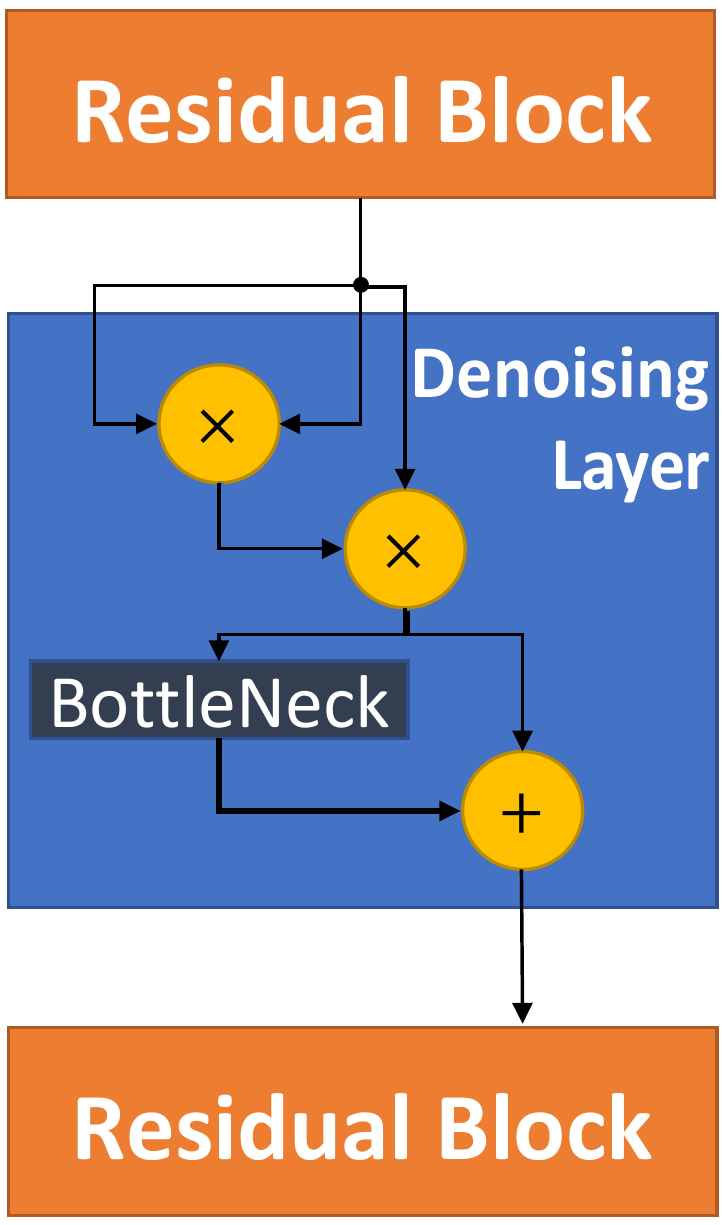}
        \caption{inner denoising layer with bottleneck}
        \label{fig:3a}
    \end{subfigure}
    \begin{subfigure}[ht]{0.33\textwidth}
        \centering
        \includegraphics[width=0.5\columnwidth,height=138pt]{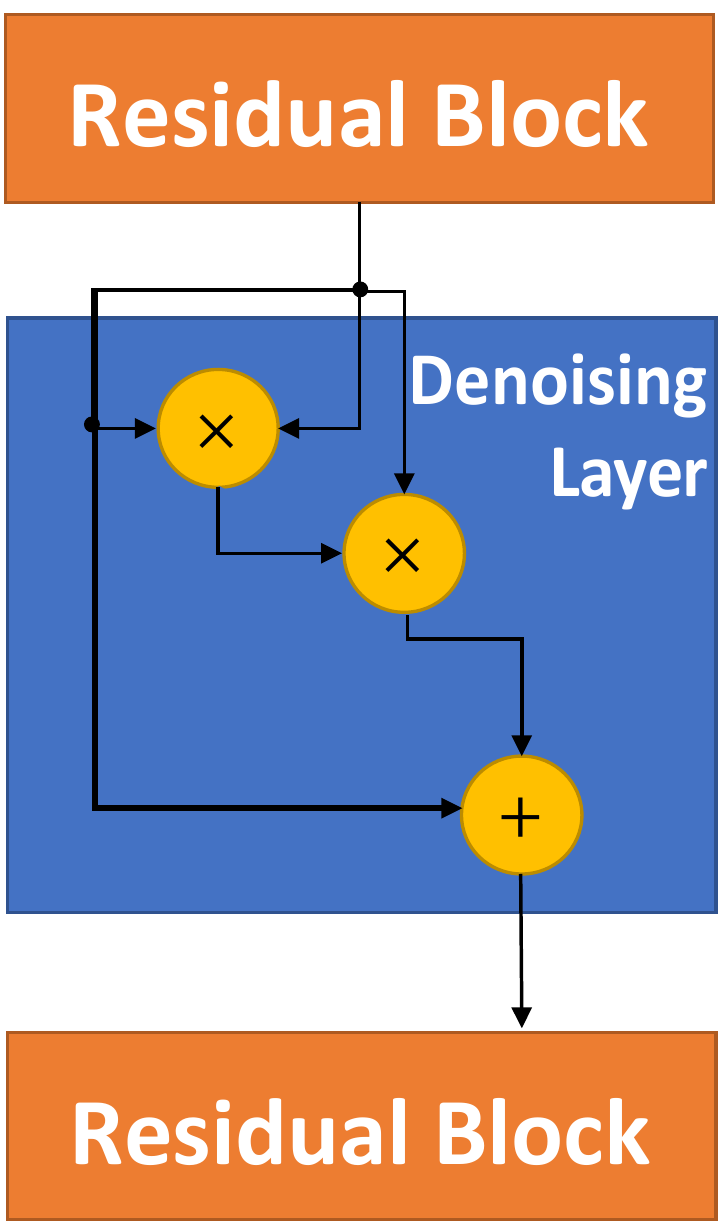}
        \caption{inner denoising layer without bottleneck}
        \label{fig:3b}
    \end{subfigure}
    \begin{subfigure}[ht]{0.33\textwidth}
        \centering
        \includegraphics[width=0.5\columnwidth,height=138pt]{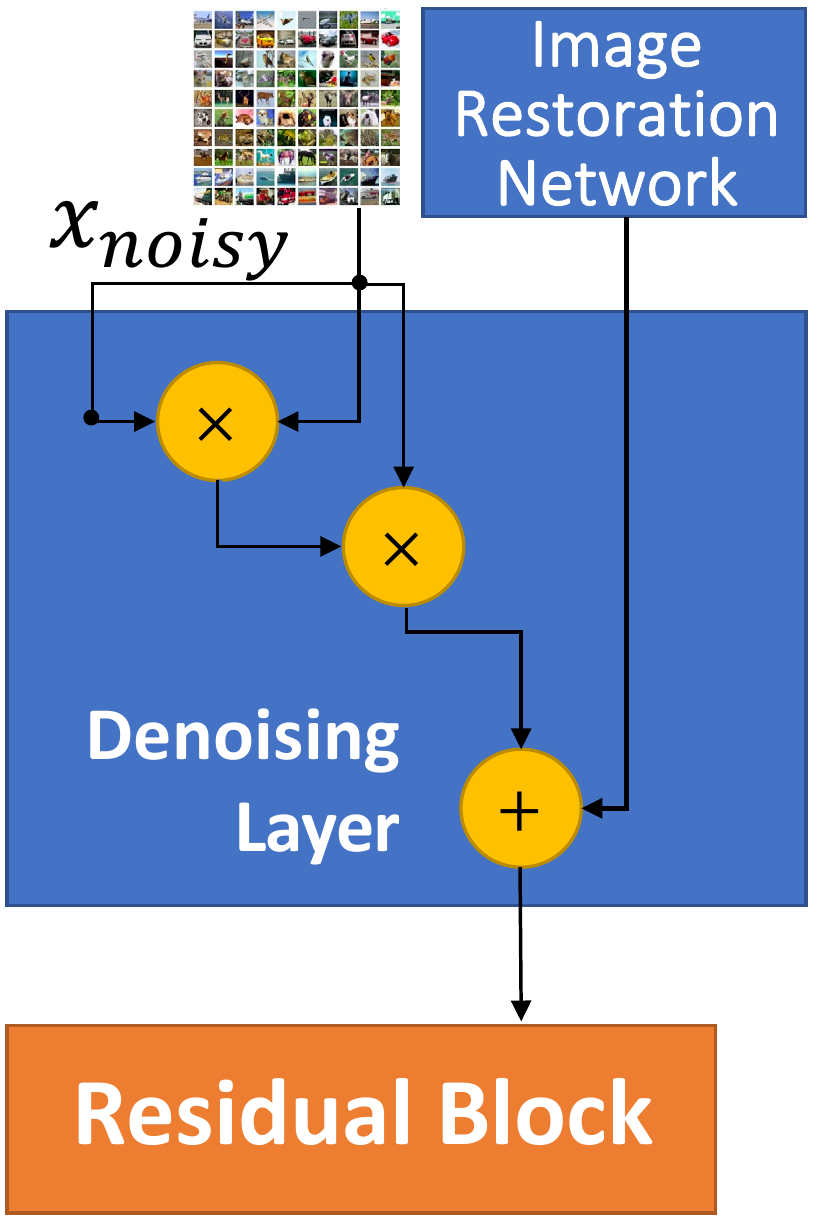}
        \caption{middle denoising layer without bottleneck}
        \label{fig:3c}
    \end{subfigure}
    \vspace{-1ex}
    \caption{Three types of denoising layers which we have experimented on. (a) an inner denoising layer that linking two residual blocks with bottleneck, (b) an inner denoising layer that linking two residual blocks without bottleneck and (c) a middle denoising layer that denoising the input of the last part without bottleneck.}
    \label{fig:3}
    \vspace{-1ex}
\end{figure*}


\section{Related Work} \label{rlwk}

\paragraph{Adversarial attack and training }
White-box attacks are typically constructed based on the gradients of the target network such as Fast-Gradient Sign Method (FGSM), Projected Gradient Descent (PGD) and Basic Iterative Method (BIM)~\cite{goodfellow_explaining_2014,madry_towards_2017,kurakin_adversarial_2016}. Some approaches focus on optimizing attack objective function like Carlini \& Wagner attack (C\&W) and DeepFool~\cite{carlini_adversarial_2017,moosavi-dezfooli_deepfool:_2016}, while others utilize the decision boundary to attack the network~\cite{brendel_decision-based_2018,chen_boundary_2019}. Black-box attacks mainly rely on transfer-attack. Attackers substitute the target network with a network,  trained with the same dataset. Subsequently, white-box attacks are applied to the substituted network for generating the adversarial samples. 

Adversarial training, proposed by~\cite{goodfellow_explaining_2014,madry_towards_2017,tramer_ensemble_2018,yan_deep_2018}, is an approach to improve the robustness of the target network. Normally, it augments the adversarial samples to the training set in the process of retraining phase. Adversarial training could achieve good results on defending against white-box and black-box attacks. However, it requires to involve a sufficient amount and variety of adversarial samples, leading to a high time-complexity. 

\vspace{-2ex}
\paragraph{Denoising}
Most denoising methods improve the intrinsic robustness of the target network, contributed by obfuscating gradients: non-differentiable operations, gradient vanishing (exploding). Various non-differentiable operations are proposed such as image quilting, total variance minimization and quantization~\cite{efros_image_2001,rudin_nonlinear_1992,guo_countering_2018}. Pixel denoising approach utilizes gradient vanishing (exploding) to thwart the attack, widely developed based on Generative-Adversarial-Network (GAN) such as~\cite{meng_magnet:_2017}. However, the aforementioned approaches cannot thwart structure-replaced white-box attacks easily~\cite{athalye_obfuscated_2018}. Attackers could still conduct attacks by approximating gradients of their non-differentiable computations. Instead of relying on obfuscating gradients, our differentiable FPD can circumvent the structure-replaced white-box attack.

Our proposal is partially related to~\cite{xie_feature_2019}, as the denoising layers in our FPD are inspired by their feature denoising approach. Nevertheless, different from \cite{xie_feature_2019}, the principle behind our FPD is to improve the intrinsic robustness, regardless of conducting adversarial training or not. Consequently, FPD includes not only two denoising modules, but also image restoration module and the Lipschitz constant constrained classification layer as well, establishing a multi-task and self-supervised training environment. Moreover, we employ denoising layers in a much more effective way: instead of implanting them to all blocks of the enhanced CNN, only shallow blocks are enhanced for maintaining high-level abstract semantic information. We will compare the performance between FPD-enhanced CNN and the CNN enhanced by~\cite{xie_feature_2019} in Section~\ref{expolaration}.


\section{Feature Pyramid Decoder}
In this section, we introduce each component of our Feature Pyramid Decoder, shown in Figure~\ref{fig:hl}. Firstly, we introduce the structure of the front denoising module $\mathrm{FPD}_\mathrm{FD}$ and back denoising module $\mathrm{FPD}_\mathrm{BD}$. Next, the structure of the image restoration module $\mathrm{FPD}_\mathrm{R}$ is depicted. Then, we modify the classification layer of the CNN by applying Lipschitz constant constraint ($\mathrm{FPD}_\mathrm{LCC}$). Finally, our two-phase training strategy is introduced, utilizing $\epsilon$-neighbourhood noisy images with multi-task and self-supervised learning.

\subsection{Front and Back Denoising Module}
A denoising module is a CNN implanted by certain inner denoising layers. Specifically, a group of inner denoising layers is only implanted into the shallow blocks of a block-based CNN. Consequently, the shallow features are processed to alleviate noise, whereas the deep features are directly decoded, helping to keep the abstract semantic information. Meanwhile, we employ a residual connection between denoised features and original features. In the light of it, most of original features could be kept and it helps to amend gradient update.

Moreover, we modify non-local means algorithm~\cite{buades_non-local_2005} by replacing the Gaussian filtering operator with a dot product operator. It could be regarded as a self-attention mechanism interpreting the relationship between pixels. Compared with the Gaussian filtering operator, the dot product operator helps improve the adversarial robustness~\cite{xie_feature_2019}. Meanwhile, as the dot product operator \textcolor{black}{does not involve} extra parameters, it contributes to relatively lower computational complexity. We explore two inner denoising structures shown in Figure~\ref{fig:3a} and Figure~\ref{fig:3b}. The corresponding performance comparison is conducted in Section~\ref{sec:denoise_select}. In our framework, the parameters of $\mathrm{FPD}_\mathrm{FD}$ and $\mathrm{FPD}_\mathrm{BD}$ are shared for shrinking the network size. The motivation of exploiting weight sharing mechanism has been explained in~\cite{kopuklu2019convolutional}: weight sharing mechanism not only reduces Memory Access Cost (MAC) but also provides more gradient updates to the reused layers from multiple parts of the network, leading to more diverse feature representations and helping $\mathrm{FPD}$ to generalize better.


\subsection{Image Restoration Module}
To build the restoration module, we firstly upsample feature maps from each block of $\mathrm{FPD}_\mathrm{FD}$ (except the first block) for the image dimension consistency and then the upsampled feature maps are fused. Finally, a group of the transposed convolutions transforms the fused feature maps into an image that has the same resolution as the input. On the other hand, we especially find that particular noise is brought by the $x_\mathrm{clean}'$. To minimize its influence, another middle denoising layer is applied to the $x_\mathrm{clean}'$, depicted in Figure~\ref{fig:3c}. Contributed by the image restoration module and the denoising module, it helps establish a two-phase training strategy. \looseness=-1

\subsection{Lipschitz Constant Constrained Classification}
The influence of employing Lipschitz constant on defending against the adversarial samples have been analyzed in~\cite{finlay_improved_2018,huster_limitations_2018}. As stated in our following Theorem~\ref{thm1}, the network could be sensitive to some perturbations if Softmax is directly used as the last layer's activation function. However, no network has ever adopted another output-layer activation function before Softmax in defending against adversarial samples so far.

\begin{figure}[ht]
\centering
\includegraphics[width=1\linewidth]{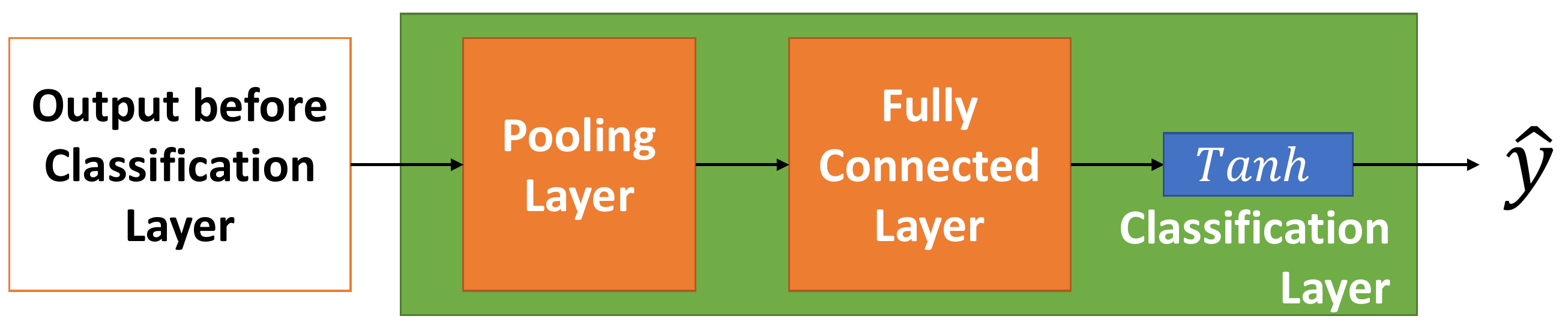}
\vspace{-1ex}
\caption{Implementation details of Lipschitz constant constrained ($\mathrm{FPD}_\mathrm{LCC}$). It is implemented by involving a squeezing activation function to the output of a fully connected layer, i.e. Tanh.}
\label{fig:classify}
\end{figure}

\begin{figure*}[ht]
\centering
\includegraphics[trim={3cm 3cm 3cm 2cm}, width=0.85\linewidth]{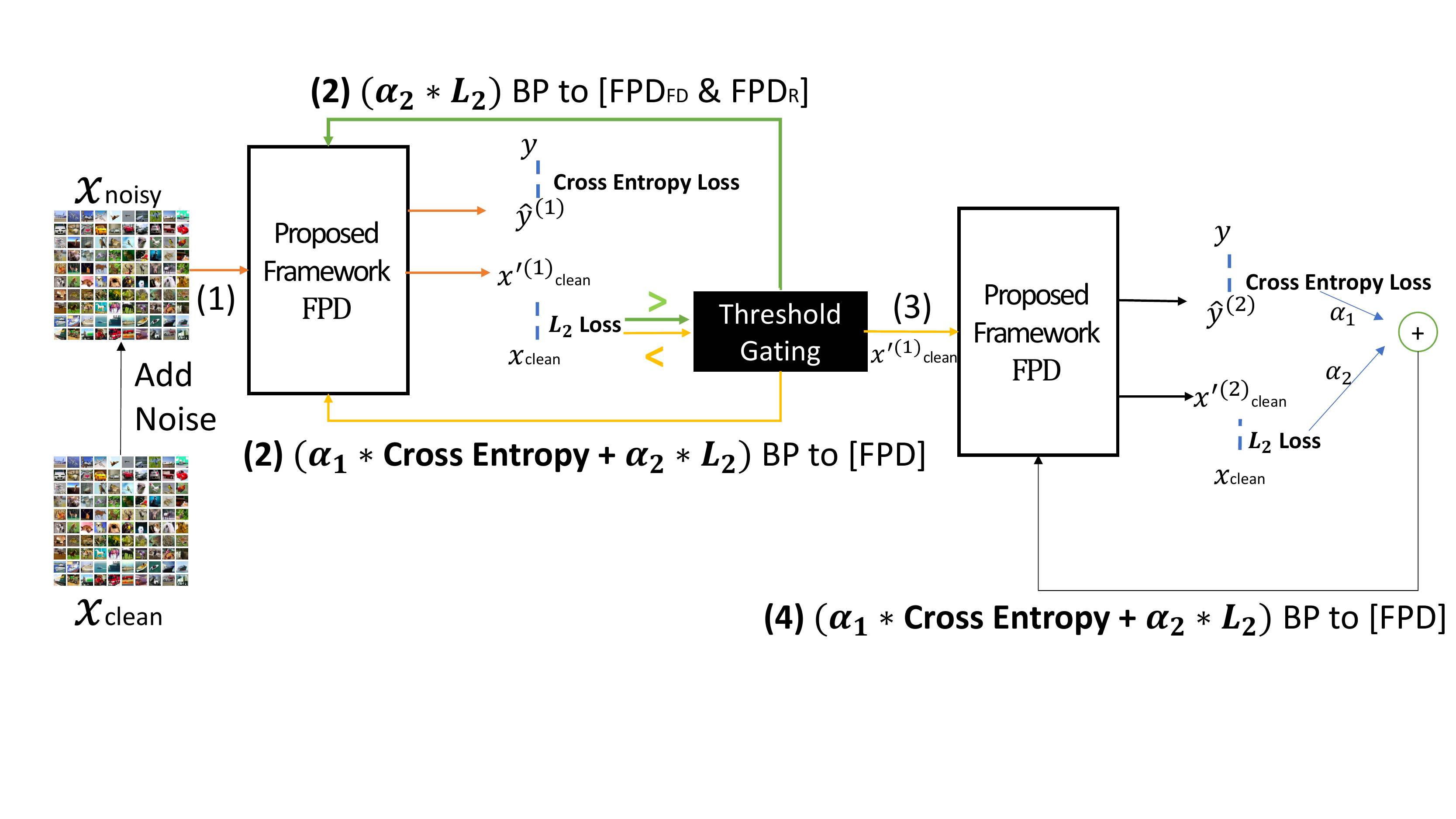}
\vspace{-1ex}
\caption{Implementation details of two-phase training strategy utilizing self-supervised and multi-task learning: the enhanced CNN FPD, in which $\mathrm{FPD}_\mathrm{R}$ refers to the image restoration module; $\mathrm{FPD}_\mathrm{FD}$ stands for the front denoising module; $\mathrm{FPD}_\mathrm{BD}$ stands for the back denoising module; $\mathrm{FPD}_\mathrm{LCC}$ refers to the modified classification layer; $x_\mathrm{noisy}$ are the samples in the $\epsilon$-neighbourhood of each image. The first phase training is optimized by $L_2(x_\mathrm{clean},x_\mathrm{clean}')$ loss. If $L_2$ loss $>T$, only the parameters of $\mathrm{FPD}_\mathrm{R}$ and $\mathrm{FPD}_\mathrm{FD}$ is updated. Once the $L_2$ loss reaches the $T$, the cross-entropy (CE) loss with $L_2$ loss jointly trains the enhanced CNN. Then, the second phase train the enhanced CNN further, jointly optimized by CE loss and $L_2$ loss. }
\label{fig:train}
\end{figure*}

\begin{thm}[the constraint on Lipschitz constant for fully-connected network]
\label{thm1}
Let $\mathrm{NN}_\mathrm{FC}$ be a $K$-way-$L$-layer-fully-connected network, $\mathrm{NN}_\mathrm{FC}(x)_{k}$ be the $k$-th component of the network output given input $x$, $w_i$ be the weight matrix of the $i$-th layer of the network, and $b_i$ be a bias matrix of the same layer. Given a noise vector $\xi$, we can bound the variation $\mathcal{V}$ component-wisely from above by:
\begin{equation*}
   \mathcal{V}_k= \left|\mathrm{NN}_\mathrm{FC}(x)_{k}-\mathrm{NN}_\mathrm{FC}(x+\xi)_{k}\right|  
 \le \frac{e^{\theta_{k}|_{x}}({e^ {\eta} }-{e^ {-\eta} })}{\sum_{p}{e^{\theta_{p}|_{x+\xi}}}}, \label{eq2}   
\end{equation*}
where
$\theta_{k}|_{x}$ is the $k$-th component of the input to {\rm Softmax} \mbox{given input $x$. Given {\rm Softmax} function as the activation func-}\newline{tion of the output layer, we denote the activation function of earlier layers by $f$, $f$'s {\rm Lipschitz constant} by $C$, and let} $\eta={\max}_{k=1,\cdots,K}\{[w_LC^{L-1}\left|w_{L-1}w_{L-2}\dots w_1\xi \right|+b_L]_k\}$.

\end{thm}

\begin{algorithm}[!ht]
\caption{Detail Training Procedures}
\label{alg:White-box}
\begin{algorithmic}[1]
\REQUIRE ~~\\
Clean images $x_\mathrm{clean}$, regenerate image $x'_\mathrm{clean}$, noisy image $x_\mathrm{noisy}$, label $y$, predict label $\hat{y}$, optimizer $opt$, updated parameters $\theta$, learning rate $lr$, weights decay $wd$, seed $s$, other hyperparameters $\alpha_{1}, \alpha_{2}, \epsilon$, the enhanced CNN FPD (including the image restoration module $\mathrm{FPD}_\mathrm{R}$, the front denoising module $\mathrm{FPD}_\mathrm{FD}$, the back denoising module $\mathrm{FPD}_\mathrm{BD}$ and the modified classification layer $\mathrm{FPD}_\mathrm{LCC}$), loss functions $L_{2}$ and cross-entropy CE, random sampler RS, threshold $T$, epoch $N$
\ENSURE ~~\\ 
FPD
\STATE Normalize each pixel of $x_\mathrm{clean}$ into $[0,1]$
\FOR{$i=1$ to $N$}
\STATE $noise:=$ RS$(s)$
\STATE UPDATE $s$
\STATE CLIP $noise$ BETWEEN $[- \epsilon ,\epsilon]$
\STATE $x_\mathrm{noisy}:=x_\mathrm{clean}+noise$
\STATE $\hat{y}^{(1)},x'^{(1)}_\mathrm{clean}:=$ FPD$(x_\mathrm{noisy})$
\STATE $l_{2}:=L_{2}(x_\mathrm{clean},\ x'^{(1)}_\mathrm{clean})$
\IF {$l_{2}>T$}
\STATE UPDATE PARAMETERS:\\ \mbox{$opt(\theta=[$$\mathrm{FPD}_\mathrm{FD},\ \mathrm{FPD}_\mathrm{R}$$],\ loss=\alpha_{2}*l_{2},\ lr,\ wd)$ }
\ELSE
\STATE $l_{1}:=$ CE$(y,\hat{y}^{(1)})$
\STATE UPDATE PARAMETERS:\\ $opt(\theta=[$FPD$],\ loss=\alpha_{1}*l_{1}+\alpha_{2}*l_{2},\ lr,\ wd)$
\STATE $\hat{y}^{(2)},x'^{(2)}_\mathrm{clean}:=$ FPD$(x'^{(1)}_\mathrm{clean})$
\STATE $l_{2}:=L_{2}(x_\mathrm{clean},\ x'^{(2)}_\mathrm{clean})$
\STATE $l_{1}:=$ CE$(y,\hat{y}^{(2)})$
\STATE UPDATE PARAMETERS:\\ $opt(\theta=[$FPD$],\ loss=\alpha_{1}*l_{1}+\alpha_{2}*l_{2},\ lr,\ wd)$
\ENDIF
\ENDFOR
\RETURN FPD
\end{algorithmic}
\end{algorithm}

We postpone the proof of Theorem~\ref{thm1} to the supplementary material. The theorem clearly shows that $w_L$ and $b_L$ may have more prominent influence than $C^{L-1}\left|w_{L-1}w_{L-2}\dots w_1\xi \right|$ on the variation of the output $\mathcal{V}_k$, when we have $0\le C\left(\sqrt[(L-1)]{\left|w_{L-1}w_{L-2}\dots w_1\xi \right|}\right)\le 1$ achieved by using regularization to restrict the weights getting close to zero. Therefore, we want to restrict $w_L$ and $b_L$ by utilizing a squeezing function $f_s$ with a small Lipschitz constant $C_s$ before Softmax in the output layer. Consequently, this reduces $\eta$ to $\max_{k=1,\dots,K}\{C_{s}C^{L-1}[\left|w_{L}w_{L-1}w_{L-2}\dots w_1\xi \right|]_k\}$, poten-\mbox{tially leading to a smaller $\mathcal{V}_k$. Therefore, the output of $\mathrm{NN}_\mathrm{FC}$} could be more stable in the face of attack perturbation.

To thwart various attacks, we let $f_s=\mathrm{Tanh}(x)$ as our squeezing function, shown in Figure~\ref{fig:classify}. Moreover, we empirically replace all the activation functions from ReLU to ELU; this leads to a smoother classification boundary, thus adapting to more complex distributions.

\begin{table}[htb]
\centering
\begin{adjustbox}{width=0.8\columnwidth,center}
\begin{tabular}{ |c||c|c|c| }
 \hline
 \multicolumn{4}{|c|}{Inner Denoising Layer Implanted Position Selection} \\
 \hline
 Accuracy & WhiteBox & BlackBox & Average\\
 \hline
 Shallow   & 1.67\%    &31.10\% & \textbf{16.39\%}\\
 Deep   & 2.08\% & 27.02\% & 14.55\%\\
 \hline
 \multicolumn{4}{|c|}{Denoising Approaches} \\
 \hline
 Average & 11.04\%    &15.99\% & 13.51\%\\
 Flip & 1.22\%    &17.34\% & 9.28\%\\
 Mid & 0.32\%    &53.77\% & \textbf{27.05\%} \\
 Mid + Inner  & 7.44\%    &42.41\% & 24.93\%\\
\hline
 \multicolumn{4}{|c|}{Ablation Study} \\
 \hline
 $\mathcal{F}_\mathrm{FD}$ & 1.67\%    &31.10\% & 16.34\%\\
 $\mathcal{F}_\mathrm{FD+R}$ & 7.44\%    &42.41\% & 24.93\%\\
 $\mathcal{F}$ & 25.55\%    & 62.72\% & \textbf{44.14\%} \\
\hline
 \multicolumn{4}{|c|}{Activation Functions Selection} \\
 \hline
 $\mathrm{ReLU}$ & 0.29\%    &49.28\% & 24.79\%\\
 $\mathrm{ELU}$ & 0.28\%    &61.24\% & 30.76\%\\
 $\mathrm{ELU+Tanh}$ & 0.25\%    & 69.11\% & \textbf{34.68\%} \\
\hline
 \multicolumn{4}{|c|}{Bottleneck Selection} \\
 \hline
 $\mathcal{F}_\mathrm{2IB-Mid}$ & 22.21\%    & 46.65\% & 34.43\%\\
 $\mathcal{F}_\mathrm{2I-Mid}$ & 25.55\%    & 62.72\% & \textbf{44.14\%}\\
\hline
 \multicolumn{4}{|c|}{No. Inner Denoising Layers Selection} \\
 \hline
 $\mathcal{F}_\mathrm{2IB}$ & 0.04\%    & 13.26\% & 6.65\%\\
 $\mathcal{F}_\mathrm{4IB}$ & 1.97\%    & 15.90\% & \textbf{8.94\%}\\
\hline
\multicolumn{4}{|c|}{Training Strategy Selection} \\
 \hline
 $\mathcal{F}_\mathrm{One\_Phase}$ & 8.60\%    & 51.08\% & 29.84\%\\
 $\mathcal{F}_\mathrm{Two\_Phase}$ & 25.55\%    & 62.72\% & \textbf{44.14\%}\\
\hline
 \multicolumn{4}{|c|}{ResNet-101 Enhanced by~\cite{xie_feature_2019}} \\
 \hline
 $\mathcal{X}$ & 5.72\%    & 62.39\% & 32.56\%\\
\hline
\end{tabular}
\end{adjustbox}
\vspace{-1ex}
\caption{Overall results of the exploration experiments with ResNet-101 on MNIST.} 
\label{tab:1}
\end{table}

\subsection{Training Strategy}
\label{sec:train}
We carefully devise our training strategy and involve uniformly sampled random noise to the clean images for further improving the enhanced CNN. Let us define the enhanced CNN FPD, in which $\mathrm{FPD}_\mathrm{R}$ refers to the image restoration module; $\mathrm{FPD}_\mathrm{FD}$ stands for the front denoising module; $\mathrm{FPD}_\mathrm{BD}$ stands for the back denoising module; $\mathrm{FPD}_\mathrm{LCC}$ refers to the modified classification layer. 

To further improve the denoising and generalization capability, we suppose that the samples in the $\epsilon$-neighbourhood of each image $x_\mathrm{clean}$ constitute adversarial samples candidate sets. We add uniformly sampled random noise to the clean images by using a sampler. It is impossible to use all samples in candidate sets, but the enhanced CNN will have more stable performance on classifying images in a smaller $\delta$-neighbourhood ($[x_\mathrm{clean}-\delta,x_\mathrm{clean}+\delta],\ 0 \le {\|\delta\|}_\infty \le \epsilon$ ) after training on noisy images. The detail training procedures are described in Algorithm~\ref{alg:White-box}. 

We propose the two-phase training to drive the self-supervised and multi-task learning for jointly optimizing the enhanced CNN. It helps the enhanced CNN to learn how to denoise images and abstract features from them with low cost and helps the enhanced CNN to learn a much more accurate mapping between images and labels. As shown in Figure~\ref{fig:train}, the first phase mainly focuses on regenerating images, optimized by $L_2(x_\mathrm{clean},x_\mathrm{clean}')$ loss. To guarantee the quality of $x_\mathrm{clean}'$ used in the later training procedures, we set a threshold $T$. If $L_2$ loss $>T$, only the parameters of $\mathrm{FPD}_\mathrm{R}$ and $\mathrm{FPD}_\mathrm{FD}$ is updated for generating the higher quality $x_\mathrm{clean}'$. Once the $L_2$ loss reaches the $T$, the cross-entropy (CE) loss with $L_2$ loss jointly trains the enhanced CNN. Then, the second phase focus on using the good quality  $x_\mathrm{clean}'$ to train the enhanced CNN further, jointly optimized by CE loss and $L_2$ loss.

\section{Experiments}
In this section, we firstly investigate the best framework structure through the exploration study. Moreover, we compare with the most related work~\cite{xie_feature_2019} as well. In the comparison experiments, we focus on comparing the robustness between the enhanced CNN and the original one, conducting adversarial training and normal training, respectively. Owing to the unavoidable limitations of evaluating robustness, we apply various attacks to evaluate our performance. However, we cannot avoid that more effective attacks exist and \mbox{the trained network will be released for future evaluation.} 

We employ MNIST, the Street View House Numbers (SVHN), \textcolor{black}{CALTECH-101 and CALTECH-256 datasets} in the following experiments. MNIST consists of a training set of 60,000 samples and a testing dataset of 10,000 samples. SVHN is a real-world colored digits image dataset. We use one of its format which includes 73,257 MNIST-like 32-by-32 images centered around a single character for training and 10,000 images for testing. For both MNIST and SVHN, we resize them to image size 64. Besides, we repeat the channel three times on MNIST for network consistency. \textcolor{black}{For both CALTECH-101 and CALTECH-256, we randomly choose 866 and 1,422 images as test images resized into 224-by-224, respectively.} We normalize image \mbox{pixel value into $[0,1]$. ResNet-101, ResNet-50~\cite{he_deep_2016} as well as} \mbox{ResNeXt-50~\cite{xie_aggregated_2017} are enhanced in the following experiments.} \mbox{We use Pytorch to implement the whole experiments.}

\subsection{Exploration Experiments}
\label{expolaration}
In this section, we conduct the exploration experiments of the FPD-enhanced CNN which is based on ResNet-101 $\mathcal{F}$ on MNIST. In Table~\ref{tab:1}, we use $L_{\infty}$-PGD attack with parameters: $\epsilon$ = 0.3, step = 40, step size = 0.01 for both white-box and black-box attacks. Under the black-box condition, we separately train a simple three layers fully-connected network as the substitute network~\cite{papernot_practical_2017} for each network. 

\vspace{-1ex}

\paragraph{Inner Denoising Layers Implanted Positions Selection}
We firstly explore the position to implant the inner denoising layers. In Table~\ref{tab:1}, 'Shallow' means that the denoising layers are implanted into the first two residual blocks. Likewise, 'Deep' means that the layers are implanted into the third and fourth residual blocks. We observe that the 'Shallow' outperforms 'Deep' on average. It may be contributed by the high-level abstract semantic information generated from the directly decoded deep features. In the following experiments, we always implant the inner denoising layers to the shallower blocks. 

\vspace{-1ex}
\paragraph{Denoising Approaches Selection}
Next, we explore the best denoising operation. In Table~\ref{tab:1}, no denoising layers are implanted in both the front and back denoising modules in 'Average', 'Flip' and 'Mid' denoising approaches. In these three approaches, we only focus on cleaning the $x'_\mathrm{clean}$ before passing to $\mathcal{F}_{\mathrm{BD}}$. Specifically, 'Average': $x_\mathrm{clean}'$ and $x_\mathrm{clean}$ are averaged; 'Flip': $x_\mathrm{clean}'$ are flipped; 'Mid': the noise in $x_\mathrm{clean}'$ are alleviated by the middle denoising layer as depicted in Figure~\ref{fig:3c}. Finally, 'Mid + Inner' means that we implant the two inner denoising layers to both the front and back denoising modules respectively. Meanwhile, the middle denoising layer is also utilized. Distinctly, 'Mid + Inner' is all-sided robust among them to defend against both the black-box and white-box attacks, attributing to the stronger denoising capability. 

\vspace{-1ex}
\paragraph{Ablation Study}
To validate the effectiveness of $\mathcal{F}$, we perform the ablation experiments on investigating the effectiveness of each module. As shown in Table~\ref{tab:1}, $\mathcal{F}$ performs far better than both $\mathcal{F}_\mathrm{FD+R}$ and $\mathcal{F}_\mathrm{FD}$ in thwarting both white-box and black-box attacks. This overall robustness is owing to the increase of data diversity and the supervision signal brought by $\mathcal{F}$. Furthermore, $\mathcal{F}_\mathrm{BD}$ can further clean the $x'_\mathrm{clean}$ to enhance the robustness in defending against the well-produced perturbations.

\vspace{-1ex}
\paragraph{\textcolor{black}{Activation Functions Selection}} \textcolor{black}{We explore the activation functions selection. Table~\ref{tab:1} indicates that ELU activation function outperforms ReLU. Furthermore, as shown in Figure~\ref{fig:classify}, ELU with Tanh achieves better performance than ELU one with 3.92\%. It demonstrates that ELU with Tanh is the suggested activation function selection.}

\begin{table*}[h]
\centering
\begin{adjustbox}{max width=1.0\textwidth}
\begin{tabular}{|c|c|c|c|c|c|c|c|c|c|c|c|c|c|c|c|c|}
\hline
                            & \multicolumn{6}{c|}{$L_{\infty}(\epsilon=0.3)$}                                  & \multicolumn{4}{c|}{$L_{2}(\epsilon=1.5)$}           & \multicolumn{4}{c|}{$L_{2}$}                              & \multicolumn{2}{c|}{}              \\ \hline
\multirow{2}{*}{Network Name} & \multicolumn{2}{c|}{FGSM} & \multicolumn{2}{c|}{PGD} & \multicolumn{2}{c|}{C\&W} & \multicolumn{2}{c|}{FGSM} & \multicolumn{2}{c|}{PGD} & \multicolumn{2}{c|}{C\&W} & \multicolumn{2}{c|}{DeepFool} & \multicolumn{2}{c|}{Average}       \\ \cline{2-17} 
                            & Acc         & T(m)        & Acc         & T(m)       & Acc          & T(m)       & Acc          & T(m)       & Acc        & T(m)        & Acc        & T(m)         & Acc            & T(m)         & Acc              & T(m)            \\ \hline
$\mathcal{O}$                         & 4\%         & 0.85        & 0\%         & 46.42      & 0\%          & 56.75      & 94\%         & 0.95       & 82\%       & 53.25       & 15\%       & 1183.67      & 6\%            & 364.13       & 27.57\%          & 243.72          \\ \hline
$\mathcal{O}_\mathrm{FGSM}$                  & 43\%        & 0.95        & 0\%         & 66.43      & 36.63\%      & 39.57      & 100\%        & 0.95       & 80\%       & 64.22       & 74\%       & 1177.27      & 6\%            & 365.17       & 48.52\%          & 244.94          \\ \hline
$\mathcal{O}_\mathrm{PGD}$                   & 92\%        & 1.85        & 76\%        & 68.37      & 89.88\%      & 42.92      & 98\%         & 1.83       & 92\%       & 68.32       & 93\%       & 1193.5       & 9\%            & 344.38       & \textbf{78.56\%} & 245.88          \\ \hline
$\mathcal{F}$                         & 31\%        & 1.5         & 0\%         & 72.3       & 88\%         & 212        & 98\%         & 1.73       & 95\%       & 98.58       & 95\%       & 1134.38      & 9.62\%         & 911.9        & \textbf{59.51\%} & \textbf{347.48} \\ \hline
$\mathcal{F}_\mathrm{FGSM}$                  & 42\%        & 0.95        & 0.87\%      & 119        & 46.12\%      & 181.6      & 97\%         & 2.25       & 95\%       & 113.97      & 97\%       & 1047.83      & 33.16\%        & 907.3        & \textbf{58.74\%} & \textbf{338.99} \\ \hline
$\mathcal{F}_\mathrm{PGD}$                   & 87\%        & 1.6         & 64.03\%     & 115.85     & 78\%         & 160        & 100\%        & 3.17       & 97\%       & 108.5       & 100\%      & 1043.25      & 11.87\%        & 912.37       & 76.84\%          & \textbf{334.96} \\ \hline
\end{tabular}
\end{adjustbox}
\vspace{-1ex}
\caption{Robustness evaluation results (Accuracy \%, Attack Time (min)) in thwarting the white-box attacks with ResNet-101 on MNIST.} 
\label{mnist-wb}
\end{table*}

\vspace{-1ex}
\paragraph{Inner Denoising Layers Selection} 
\label{sec:denoise_select}
We also investigate the optimal number of the inner denoising layers and whether to use the bottleneck in these inner layers. In Table~\ref{tab:1}, $\mathcal{F}_\mathrm{kIB-Mid}$: $\mathrm{k}$ inner denoising layers with the bottleneck as depicted in Figure~\ref{fig:3a} are implanted to each denoise module $\mathcal{F}_\mathrm{FD}$ and $\mathcal{F}_\mathrm{BD}$ respectively. Meanwhile, the middle denoising layer is used as depicted above; $\mathcal{F}_\mathrm{kIB}$ is similar to $\mathcal{F}_\mathrm{kIB-Mid}$ except that no middle denoising layer is involved in the framework. $\mathcal{F}_\mathrm{kI-Mid}$ means that the bottleneck is not used in the inner denoising layers as depicted in Figure~\ref{fig:3b}. We observe that the bottleneck reduces the performance around 10\%. Moreover, although $\mathcal{F}_\mathrm{4IB}$ outperforms $\mathcal{F}_\mathrm{2IB}$, the enhancement is not worthy if we consider the time complexity brought by the additional denoising layers. Therefore, we use $\mathcal{F}_\mathrm{2I-Mid}$ as our proposed framework in the following experiments. 

\vspace{-1ex}
\paragraph{\textcolor{black}{Training Strategy Selection}} \textcolor{black}{We further demonstrate the efficacy of our two-phase training strategy $\mathcal{F}_{\mathrm{Two\_Phase}}$ as depicts in Figure~\ref{fig:train}. We mainly compare $\mathcal{F}_{\mathrm{Two\_Phase}}$ with one-phase training strategy $\mathcal{F}_{\mathrm{One\_Phase}}$ .i.e the first training phase (describes in Section~\ref{sec:train}). Results show that $\mathcal{F}_\mathrm{Two\_Phase}$ could achieve higher performance than $\mathcal{F}_\mathrm{One\_Phase}$ with 14.3\%.}

\vspace{-1ex}
\paragraph{Comparison with the Related Work}
As mentioned in Section~\ref{rlwk}, the denoising approach proposed in~\cite{xie_feature_2019} is similar to our denoising layers in FPD. Therefore, we conduct a comparison experiment with~\cite{xie_feature_2019} as well. In Table~\ref{tab:1}, $\mathcal{X}$ represents the enhanced CNN by~\cite{xie_feature_2019}. We observe that our $\mathcal{F}_\mathrm{2I-Mid}$ outperforms $\mathcal{X}$. Especially, the performance of thwarting the white-box attack is about 20\% higher.

\subsection{Comparison Experiments}

We conduct a series of comparison experiments\footnote{We use adversarial-robustness-toolbox~\cite{nicolae_adversarial_2018}, a tool for testing the network's robustness of defending against various attacks.} to further evaluate FPD-enhanced CNN performance on MNIST, SVHN, CALTECH-101 and CALTECH-256.
\vspace{-5pt}
\paragraph{Notation and Implementation Details} Firstly, let us define the following notations for accurate description:  $\mathcal{F}$ represents the enhanced CNN; $\mathcal{O}$ is the original CNN; $\mathcal{F}_\mathrm{PGD}$ and $\mathcal{O}_\mathrm{PGD}$ is adversarial trained by $L_\infty$-PGD (on MNIST: $\epsilon$=0.3, step=100 and step length=0.01; on SVHN: $\epsilon$=8/256.0, step=40 and step length=2/256.0); $\mathcal{F}_\mathrm{FGSM}$ and $\mathcal{O}_\mathrm{FGSM}$ is adversarial trained by $L_\infty$-FGSM (on MNIST: $\epsilon$=0.3). All results are achieved with the batch size 100, running on the RTX Titan.

\vspace{-5pt}
\paragraph{On MNIST} For sufficient evaluation, we firstly focus on applying FPD to ResNet-101 on MNIST. We mainly concentrate on two performance metrics: classification accuracy and attack time. Longer attack time can be a result of a monetary limit. In this perspective, we believe that longer attacking time may result in the excess of time and monetary limit. The attacker may surrender the attack. Therefore, we state that attackers spend more time attacking networks, which may protect the networks from another perspective.


We employ various white-box attacks to attack $\mathcal{F}$, $\mathcal{O}$, $\mathcal{F}_\mathrm{PGD}$, $\mathcal{O}_\mathrm{PGD}$, $\mathcal{F}_\mathrm{FGSM}$ and $\mathcal{O}_\mathrm{FGSM}$. We consider following attacks, including $L_2$-PGD, $L_2$-FGSM, $L_\infty$-PGD and $L_\infty$-FGSM. We set $\epsilon$=1.5 and 0.3 to bound the permutations for $L_2$ and $L_\infty$ norm. Both $L_2$-PGD and $L_\infty$-PGD are set to attack for 100 iterations and each step length is 0.1.

\begin{table*}[h]
\centering
\begin{adjustbox}{max width=1\textwidth}
\begin{tabular}{|c|c|c|c|l|c|l|c|l|c|l|c|l|c|l|c|l|c|l|c|l|c|l|c|l|c|l|c|l|c|l|c|l|c|}
\hline
\multicolumn{2}{|c|}{\multirow{5}{*}{Network Name}} & \multirow{4}{*}{\begin{tabular}[c]{@{}c@{}}Clean\\ Examples\end{tabular}} & \multicolumn{31}{c|}{BlackBox}                                                                                                                                                                                                                                                                                                                                                                                                                                                                \\ \cline{4-34} 
\multicolumn{2}{|c|}{}                            &                                 & \multicolumn{10}{c|}{ResNet-101}                                                                                                                        & \multicolumn{10}{c|}{ResNet-50}                                                                                                                         & \multicolumn{10}{c|}{ResNeXt-50}                                                                                                                      & \multirow{2}{*}{} \\ \cline{4-33}
\multicolumn{2}{|c|}{}                            &                                 & \multicolumn{6}{c|}{Substitute:$\mathcal{O}$}                                                        & \multicolumn{4}{c|}{Substitute:$\mathcal{F}$}                        & \multicolumn{6}{c|}{Substitute:$\mathcal{O}$}                                                        & \multicolumn{4}{c|}{Substitute:$\mathcal{F}$}                        & \multicolumn{6}{c|}{Substitute:$\mathcal{O}$}                                                     & \multicolumn{4}{c|}{Substitute:$\mathcal{F}$}                         &                   \\ \cline{4-34} 
\multicolumn{2}{|c|}{}                            &                                 & \multicolumn{2}{c|}{FGSM}    & \multicolumn{2}{c|}{PGD}     & \multicolumn{2}{c|}{C\&W}    & \multicolumn{2}{c|}{FGSM}   & \multicolumn{2}{c|}{PGD}     & \multicolumn{2}{c|}{FGSM}    & \multicolumn{2}{c|}{PGD}     & \multicolumn{2}{c|}{C\&W}    & \multicolumn{2}{c|}{FGSM}   & \multicolumn{2}{c|}{PGD}     & \multicolumn{2}{c|}{FGSM} & \multicolumn{2}{c|}{PGD}     & \multicolumn{2}{c|}{C\&W}    & \multicolumn{2}{c|}{FGSM}    & \multicolumn{2}{c|}{PGD}     & Average           \\ \cline{3-34} 
\multicolumn{2}{|c|}{}                            & Acc                             & \multicolumn{2}{c|}{Acc}     & \multicolumn{2}{c|}{Acc}     & \multicolumn{2}{c|}{Acc}     & \multicolumn{2}{c|}{Acc}    & \multicolumn{2}{c|}{Acc}     & \multicolumn{2}{c|}{Acc}     & \multicolumn{2}{c|}{Acc}     & \multicolumn{2}{c|}{Acc}     & \multicolumn{2}{c|}{Acc}    & \multicolumn{2}{c|}{Acc}     & \multicolumn{2}{c|}{Acc}  & \multicolumn{2}{c|}{Acc}     & \multicolumn{2}{c|}{Acc}     & \multicolumn{2}{c|}{Acc}     & \multicolumn{2}{c|}{Acc}     & Acc               \\ \hline
\multirow{2}{*}{ResNet-101}      & $\mathcal{O}_\mathrm{PGD}$      & 89\%                            & \multicolumn{2}{c|}{87\%}    & \multicolumn{2}{c|}{87\%}    & \multicolumn{2}{c|}{86\%}    & \multicolumn{2}{c|}{80\%}   & \multicolumn{2}{c|}{86\%}    & \multicolumn{2}{c|}{86\%}    & \multicolumn{2}{c|}{87\%}    & \multicolumn{2}{c|}{86\%}    & \multicolumn{2}{c|}{81\%}   & \multicolumn{2}{c|}{88.01\%} & \multicolumn{2}{c|}{84\%} & \multicolumn{2}{c|}{84\%}    & \multicolumn{2}{c|}{86\%}    & \multicolumn{2}{c|}{92\%}    & \multicolumn{2}{c|}{85.12\%} & \textbf{85.68\%}  \\ \cline{2-34} 
                                 & $\mathcal{F}_\mathrm{PGD}$      & 84\%                            & \multicolumn{2}{c|}{82\%}    & \multicolumn{2}{c|}{83\%}    & \multicolumn{2}{c|}{84\%}    & \multicolumn{2}{c|}{64\%}   & \multicolumn{2}{c|}{78\%}    & \multicolumn{2}{c|}{82\%}    & \multicolumn{2}{c|}{83\%}    & \multicolumn{2}{c|}{84\%}    & \multicolumn{2}{c|}{76\%}   & \multicolumn{2}{c|}{83\%}    & \multicolumn{2}{c|}{80\%} & \multicolumn{2}{c|}{82\%}    & \multicolumn{2}{c|}{84\%}    & \multicolumn{2}{c|}{82\%}    & \multicolumn{2}{c|}{82\%}    & 80.6\%            \\ \hline
\multirow{2}{*}{ResNet-50}       & $\mathcal{O}_\mathrm{PGD}$      & 85\%                            & \multicolumn{2}{c|}{81\%}    & \multicolumn{2}{c|}{83\%}    & \multicolumn{2}{c|}{88\%}    & \multicolumn{2}{c|}{78\%}   & \multicolumn{2}{c|}{81\%}    & \multicolumn{2}{c|}{80\%}    & \multicolumn{2}{c|}{82\%}    & \multicolumn{2}{c|}{88\%}    & \multicolumn{2}{c|}{76\%}   & \multicolumn{2}{c|}{80\%}    & \multicolumn{2}{c|}{92\%} & \multicolumn{2}{c|}{92\%}    & \multicolumn{2}{c|}{88\%}    & \multicolumn{2}{c|}{92\%}    & \multicolumn{2}{c|}{92\%}    & \textbf{84.87\%}  \\ \cline{2-34} 
                                 & $\mathcal{F}_\mathrm{PGD}$      & 89\%                            & \multicolumn{2}{c|}{87\%}    & \multicolumn{2}{c|}{88\%}    & \multicolumn{2}{c|}{89\%}    & \multicolumn{2}{c|}{77\%}   & \multicolumn{2}{c|}{87\%}    & \multicolumn{2}{c|}{87\%}    & \multicolumn{2}{c|}{88\%}    & \multicolumn{2}{c|}{89\%}    & \multicolumn{2}{c|}{62\%}   & \multicolumn{2}{c|}{71\%}    & \multicolumn{2}{c|}{86\%} & \multicolumn{2}{c|}{90\%}    & \multicolumn{2}{c|}{88\%}    & \multicolumn{2}{c|}{92\%}    & \multicolumn{2}{c|}{92\%}    & \textbf{84.87\%}  \\ \hline
\multirow{2}{*}{ResNeXt-50}      & $\mathcal{O}_\mathrm{PGD}$     & 96\%                            & \multicolumn{2}{c|}{92\%}    & \multicolumn{2}{c|}{93.48\%} & \multicolumn{2}{c|}{96\%}    & \multicolumn{2}{c|}{86\%}   & \multicolumn{2}{c|}{93.45\%} & \multicolumn{2}{c|}{92\%}    & \multicolumn{2}{c|}{94.97\%} & \multicolumn{2}{c|}{96\%}    & \multicolumn{2}{c|}{90\%}   & \multicolumn{2}{c|}{92\%}    & \multicolumn{2}{c|}{84\%} & \multicolumn{2}{c|}{86\%}    & \multicolumn{2}{c|}{92\%}    & \multicolumn{2}{c|}{92\%}    & \multicolumn{2}{c|}{94\%}    & \textbf{91.59\%}  \\ \cline{2-34} 
                                 & $\mathcal{F}_\mathrm{PGD}$      & 86\%                            & \multicolumn{2}{c|}{80\%}    & \multicolumn{2}{c|}{84\%}    & \multicolumn{2}{c|}{86\%}    & \multicolumn{2}{c|}{74\%}   & \multicolumn{2}{c|}{82\%}    & \multicolumn{2}{c|}{84\%}    & \multicolumn{2}{c|}{84.81\%} & \multicolumn{2}{c|}{86\%}    & \multicolumn{2}{c|}{80\%}   & \multicolumn{2}{c|}{84\%}    & \multicolumn{2}{c|}{66\%} & \multicolumn{2}{c|}{62\%}    & \multicolumn{2}{c|}{86\%}    & \multicolumn{2}{c|}{80\%}    & \multicolumn{2}{c|}{82\%}    & 80.05\%           \\ \hline
\multicolumn{2}{|c|}{Average Acc}                 & 88.17\%                         & \multicolumn{2}{c|}{84.83\%} & \multicolumn{2}{c|}{86.41\%} & \multicolumn{2}{c|}{88.17\%} & \multicolumn{2}{c|}{76.5\%} & \multicolumn{2}{c|}{84.58\%} & \multicolumn{2}{c|}{85.17\%} & \multicolumn{2}{c|}{86.63\%} & \multicolumn{2}{c|}{88.17\%} & \multicolumn{2}{c|}{77.5\%} & \multicolumn{2}{c|}{83\%}    & \multicolumn{2}{c|}{82\%} & \multicolumn{2}{c|}{82.67\%} & \multicolumn{2}{c|}{87.33\%} & \multicolumn{2}{c|}{88.33\%} & \multicolumn{2}{c|}{87.85\%} & 84.61\%           \\ \hline
\end{tabular}
\end{adjustbox}
\vspace{-1ex}
\caption{$L_\infty$ Metrics: Robustness evaluation results (Accuracy \%) in thwarting the black-box attacks with ResNet-101, ResNet-50 and ResNeXt-50 on SVHN.}
\label{sblk}
\end{table*}

\begin{table}[h]
\begin{adjustbox}{max width=1.05\columnwidth}
\begin{tabular}{|c|c|c|l|c|l|c|l|c|l|c|l|c|}
\hline
\multicolumn{2}{|c|}{\multirow{4}{*}{Network Name}} & \multicolumn{11}{c|}{WhiteBox}                                                                                                                                         \\ \cline{3-13} 
\multicolumn{2}{|c|}{}                            & \multicolumn{6}{c|}{$L_{\infty}(\epsilon=8/256.0)$}                                  & \multicolumn{4}{c|}{$L_{2}$}                                 &                  \\ \cline{3-13} 
\multicolumn{2}{|c|}{}                            & \multicolumn{2}{c|}{FGSM} & \multicolumn{2}{c|}{PGD}  & \multicolumn{2}{c|}{C\&W}    & \multicolumn{2}{c|}{C\&W}    & \multicolumn{2}{c|}{DeepFool} & Average          \\ \cline{3-13} 
\multicolumn{2}{|c|}{}                            & \multicolumn{2}{c|}{Acc}  & \multicolumn{2}{c|}{Acc}  & \multicolumn{2}{c|}{Acc}     & \multicolumn{2}{c|}{Acc}     & \multicolumn{2}{c|}{Acc}      & Acc              \\ \hline
\multirow{4}{*}{ResNet-101}      & $\mathcal{O}$           & \multicolumn{2}{c|}{1\%}  & \multicolumn{2}{c|}{0\%}  & \multicolumn{2}{c|}{0\%}     & \multicolumn{2}{c|}{0\%}     & \multicolumn{2}{c|}{28\%}     & 5.8\%            \\ \cline{2-13} 
                                 & $\mathcal{O}_\mathrm{PGD}$      & \multicolumn{2}{c|}{57\%} & \multicolumn{2}{c|}{36\%} & \multicolumn{2}{c|}{39\%}    & \multicolumn{2}{c|}{1\%}     & \multicolumn{2}{c|}{3\%}      & 27.2\%  \\ \cline{2-13} 
                                 & $\mathcal{F}$            & \multicolumn{2}{c|}{44\%} & \multicolumn{2}{c|}{44\%} & \multicolumn{2}{c|}{71\%}    & \multicolumn{2}{c|}{62.22\%} & \multicolumn{2}{c|}{53.25\%}  & 54.89\%          \\ \cline{2-13} 
                                 & $\mathcal{F}_\mathrm{PGD}$      & \multicolumn{2}{c|}{48\%} & \multicolumn{2}{c|}{47\%} & \multicolumn{2}{c|}{72.57\%} & \multicolumn{2}{c|}{77.7\%}  & \multicolumn{2}{c|}{57\%}     & \textbf{60.45\%} \\ \hline
\multirow{4}{*}{ResNet-50}       & $\mathcal{O}$            & \multicolumn{2}{c|}{4\%}  & \multicolumn{2}{c|}{0\%}  & \multicolumn{2}{c|}{0\%}     & \multicolumn{2}{c|}{0\%}     & \multicolumn{2}{c|}{34\%}     & 7.6\%            \\ \cline{2-13} 
                                 & $\mathcal{O}_\mathrm{PGD}$      & \multicolumn{2}{c|}{55\%} & \multicolumn{2}{c|}{26\%} & \multicolumn{2}{c|}{28\%}    & \multicolumn{2}{c|}{0\%}     & \multicolumn{2}{c|}{11\%}     & 24\%    \\ \cline{2-13} 
                                 & $\mathcal{F}$            & \multicolumn{2}{c|}{33\%} & \multicolumn{2}{c|}{30\%} & \multicolumn{2}{c|}{61\%}    & \multicolumn{2}{c|}{52.03\%} & \multicolumn{2}{c|}{36.78\%}  & 42.56\%          \\ \cline{2-13} 
                                 & $\mathcal{F}_\mathrm{PGD}$      & \multicolumn{2}{c|}{39\%} & \multicolumn{2}{c|}{35\%} & \multicolumn{2}{c|}{70\%}    & \multicolumn{2}{c|}{73.3\%}  & \multicolumn{2}{c|}{45.38\%}  & \textbf{52.54\%} \\ \hline
\multirow{4}{*}{ResNeXt-50}      & $\mathcal{O}$            & \multicolumn{2}{c|}{13\%} & \multicolumn{2}{c|}{0\%}  & \multicolumn{2}{c|}{0\%}     & \multicolumn{2}{c|}{0.4\%}   & \multicolumn{2}{c|}{51.24\%}  & 12.93\%          \\ \cline{2-13} 
                                 & $\mathcal{O}_\mathrm{PGD}$      & \multicolumn{2}{c|}{58\%} & \multicolumn{2}{c|}{36\%} & \multicolumn{2}{c|}{46\%}    & \multicolumn{2}{c|}{4.5\%}   & \multicolumn{2}{c|}{24.27\%}  & 33.75\% \\ \cline{2-13} 
                                 & $\mathcal{F}$            & \multicolumn{2}{c|}{80\%} & \multicolumn{2}{c|}{80\%} & \multicolumn{2}{c|}{86\%}    & \multicolumn{2}{c|}{86\%}    & \multicolumn{2}{c|}{83.17\%}  & \textbf{83.03\%} \\ \cline{2-13} 
                                 & $\mathcal{F}_\mathrm{PGD}$      & \multicolumn{2}{c|}{80\%} & \multicolumn{2}{c|}{78\%} & \multicolumn{2}{c|}{86\%}    & \multicolumn{2}{c|}{84.15\%} & \multicolumn{2}{c|}{84\%}     & 82.43\%          \\ \hline
\end{tabular}
\end{adjustbox}
\vspace{-1ex}
\caption{Robustness evaluation results (Accuracy \%) in thwarting the white-box attacks with ResNet-101, ResNet-50 and ResNeXt-50 on SVHN.}
\label{sw}
\end{table}

We have the following remarks on our results as shown in Table~\ref{mnist-wb}. Generally, $\mathcal{F}$ and its adversarial trained $\mathcal{F}_\mathrm{FGSM}$ outperform $\mathcal{O}$ and $\mathcal{O}_\mathrm{FGSM}$ in accuracy around 32\% and 10\%, respectively. $\mathcal{O}_\mathrm{PGD}$ seems slightly more robust than $\mathcal{F}_\mathrm{PGD}$. However, as revealed by the average attack time, more computational time (around 89 min) is spent on attacking $\mathcal{F}_\mathrm{PGD}$. In particular, the overall time spent on attacking $\mathcal{F}$, its adversarial trained networks $\mathcal{F}_\mathrm{FGSM}$, $\mathcal{F}_\mathrm{PGD}$ are longer than $\mathcal{O}$, $\mathcal{O}_\mathrm{FGSM}$ and $\mathcal{O}_\mathrm{PGD}$ around 104 min, \mbox{94 min and 89 min. Above results have demonstrated that $\mathcal{F}$} \mbox{and its adversarial trained networks are harder to be attacked.}

\vspace{-5pt}
\paragraph{On SVHN}

We mainly assess the ability of FPD to enhance various block-based CNNs on colored samples: ResNet-101, ResNet-50, ResNeXt-50. We employ a series of white-box and black-box attacks to attack $\mathcal{F}$, $\mathcal{O}$, $\mathcal{F}_\mathrm{PGD}$ and $\mathcal{O}_\mathrm{PGD}$ for each block-based CNNs. Initially, we evaluate FPD performance in thwarting black-box attacks. As shown in Table ~\ref{sblk}, $\mathcal{O}$ and $\mathcal{F}$ of each block-based CNNs are employed as substitutes. We adopt $L_\infty$-FGSM and $L_\infty$-PGD to attack them. Besides, we observe that $O$ is hard to defend against a $L_\infty$-C\&W attack, depicted in Table~\ref{sw}. Therefore, we additionally adopt $L_\infty$-C\&W to attack substitute $\mathcal{O}$, to further evaluate FPD. As for white-box attacks, we adopt following attacks: $L_\infty$-FGSM, $L_\infty$-PGD, $L_\infty$-C\&W, $L_2$-DeepFool and $L_2$-C\&W. We set $\epsilon$=8/256.0 for above-mentioned attacks and PGD is set to attack for 40 iterations with step length 2/256.0.

\begin{figure}[h]
    \centering
    \begin{subfigure}[ht]{0.325\columnwidth}
        \centering
        \includegraphics[width = 0.9\columnwidth]{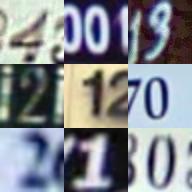}
        \caption{Adversarial \protect\\ images.}
        \label{fig:adv}
    \end{subfigure}
    \begin{subfigure}[ht]{0.325\columnwidth}
        \centering
        \includegraphics[width = 0.9\columnwidth]{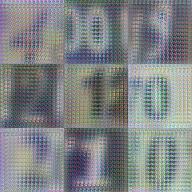}
        \caption{Restoration \protect\\ (Adv).}
        \label{fig:res_adv}
    \end{subfigure}
    \begin{subfigure}[ht]{0.325\columnwidth}
        \centering
        \includegraphics[width = 0.9\columnwidth]{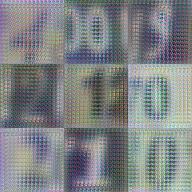}
        \caption{Restoration \protect\\ (Clean).}
        \label{fig:res_clean}
    \end{subfigure}
    \vspace{-1ex}
    \caption{Adversarial images (a) vs. the output of image restoration module from adversarial images (b) and clean images (c). \textcolor{black}{Images are reproduced from the data in Table \ref{sw} (enhanced ResNet-101 attacked by PGD).}}
    \label{fig:res}
\end{figure}

We have the following remarks on our results as shown in Table~\ref{sblk} and Table~\ref{sw}. Firstly, in defending against white-box attacks, $\mathcal{F}$ and the adversarial trained $\mathcal{F}_\mathrm{PGD}$ far outperform $\mathcal{O}$ and $\mathcal{O}_\mathrm{PGD}$ in accuracy for all the block-based CNNs, especially in ResNet-101 and ResNeXt-50. We notice that the performance of $\mathcal{F}_\mathrm{PGD}$ is not exactly satisfactory under black-box attacks, yet the outcome is not very surprising. As shown in Table~\ref{sw}, $\mathcal{F}$-based networks achieve a high accuracy under white-box attacks. Therefore, when these attacks are applied to $\mathcal{F}$ substitute, some attacks effectively fail, returning a large number of clean samples as adversarial examples. Given that $\mathcal{F}_\mathrm{PGD}$ has a lower accuracy than $\mathcal{O}_\mathrm{PGD}$ on clean samples for ResNet-101 and ResNeXt-50 (as depicted in Table~\ref{sblk}), $\mathcal{F}$-based networks achieve this biased performance under black-box attacks. 

We also show the output of image restoration module in Figure~\ref{fig:res}. \textcolor{black}{Adversarial images are well ``denoised'' by comparing Figure~\ref{fig:adv} with~\ref{fig:res_adv}}.  Figure~\ref{fig:res_adv} and~\ref{fig:res_clean} \textcolor{black}{illustrate that} the module output generated by adversarial and clean images are quite \textcolor{black}{similar}. It guarantees that restoration module could generate similar images from both adversarial and clean images for $\mathrm{FPD}_\mathrm{BD}$, leading to more robust performance in defending against attacks.

\vspace{-5pt}
\paragraph{On CALTECH-101 \& CALTECH-256}


\textcolor{black}{We further demonstrate the efficacy of FPD on ResNet-101 on high dimensional dataset CALTECH-101 and CALTECH-256, attacked by $L_{\infty}$-PGD attack for 40 iterations. For this attack, we set $\epsilon$ to 8/256.0 and step length to 2/256.0. To be specific, on CALTECH-101, $\mathcal{F}$ achieves 61.78\% under PGD attack. It outperforms $\mathcal{O}$ around 34.64\%. On CALTECH-256, our ResNet-101 model $\mathcal{F}$ achieve 49.79\% accuracy against 0.00\% of the original one $\mathcal{O}$.}

In summary, above results have demonstrated that the FPD-enhanced CNN is much more robust than non-enhanced versions on MNIST and high dimensional dataset CALTECH-101 and CALTECH-256. On colored dataset SVHN, the performance under black-box attacks is not exactly satisfactory. However, considering the performance in thwarting white-box attacks, FPD-enhanced CNN performs far better than non-enhanced versions.



\section{Conclusion}
In this paper, we have presented a novel Feature Pyramid Decoder (FPD) to enhance the intrinsic robustness of the block-based CNN. Besides, we have devised a novel two-phase training strategy. Through the exploration experiments, we have investigated the best structure of our FPD. Moreover, we go through a series of comparison experiments to demonstrate the effectiveness of the FPD. Attacking these models by a variety of white-box and black-box attacks, we have shown that the proposed FPD can enhance the robustness of the CNNs. We are planning to design a more powerful decoder to improve desnoising capability. Also, we will exploit a hard threshold to filter relatively bad restored images, further improving classification accuracy. Finally, we will transplant FPD to non-block CNN.

\section{Acknowledgement}
This paper is supported by the Fundamental Research Fund of Shandong Academy of Sciences (NO. 2018:12-16), Major Scientific and Technological Innovation Projects of Shandong Province, China (No. 2019JZZY020128), as well as AcRF Tier 2 Grant MOE2016-T2-2-022 and AcRF Tier 1 Grant RG17/19, Singapore. 
\vfill
\eject

{\small
\bibliographystyle{ieee_fullname}
\bibliography{bib}

\begin{thebibliography}{10}\itemsep=-1pt

\bibitem{athalye_obfuscated_2018}
Anish Athalye, Nicholas Carlini, and David~A. Wagner.
\newblock {{Obfuscated {Gradients} {Give} a {False} {Sense} of {Security}:
  {Circumventing} {Defenses} to {Adversarial} {Examples}}}.
\newblock {\em CoRR}, abs/1802.00420, 2018.

\bibitem{brendel_decision-based_2018}
Wieland Brendel, Jonas Rauber, and Matthias Bethge.
\newblock {{Decision-{Based} {Adversarial} {Attacks}: {Reliable} {Attacks}
  {against} {Black}-{Box} {Machine} {Learning} {Models}}}.
\newblock In {\em Proc. of the ICLR}, 2018.

\bibitem{buades_non-local_2005}
Antoni Buades, Bartomeu Coll, and Jean-Michel Morel.
\newblock {A {Non}-{Local} {Algorithm} for {Image} {Denoising}}.
\newblock In {\em Proc. of the CVPR}, pages 60--65, 2005.

\bibitem{carlini_towards_2017}
Nicholas Carlini and David Wagner.
\newblock {{Towards Evaluating the Robustness of Neural Networks}}.
\newblock In {\em Symposium on {Security} and {Privacy} ({SP})}, pages 39--57,
  2017.

\bibitem{carlini_adversarial_2017}
Nicholas Carlini and David~A. Wagner.
\newblock {{Adversarial {Examples} {Are} {Not} {Easily} {Detected}: {Bypassing}
  {Ten} {Detection} {Methods}}}.
\newblock {\em CoRR}, abs/1705.07263, 2017.

\bibitem{che_adversarial_2019}
Zhaohui Che, Ali Borji, Guangtao Zhai, Suiyi Ling, Guodong Guo, and Patrick
  Le~Callet.
\newblock {{Adversarial {Attacks} against {Deep} {Saliency} {Models}}}.
\newblock {\em CoRR}, abs/1904.01231, 2019.

\bibitem{chen_boundary_2019}
Jianbo Chen and Michael~I Jordan.
\newblock {{Boundary Attack++: {Query}-efficient Decision-based Adversarial
  Attack}}.
\newblock {\em CoRR}, abs/1904.02144, 2019.

\bibitem{efros_image_2001}
Alexei~A. Efros and William~T. Freeman.
\newblock {{Image Quilting for Texture Synthesis and Transfer}}.
\newblock In {\em Proc. of the SIGGRAPH}, pages 341--346, 2001.

\bibitem{finlay_improved_2018}
Chris Finlay, Adam~M. Oberman, and Bilal Abbasi.
\newblock {{Improved Robustness to Adversarial Examples using {Lipschitz}
  Regularization of the Loss}}.
\newblock {\em CoRR}, abs/1810.00953, 2018.

\bibitem{goodfellow_explaining_2014}
Ian~J. Goodfellow, Jonathon Shlens, and Christian Szegedy.
\newblock {{Explaining and {Harnessing} {Adversarial} {Examples}}}.
\newblock {\em CoRR}, abs/1412.6572, 2014.

\bibitem{guo_countering_2018}
Chuan Guo, Mayank Rana, Moustapha Cisse, and Laurens van~der Maaten.
\newblock {{Countering {Adversarial} {Images} using {Input}
  {Transformations}}}.
\newblock In {\em Proc. of the ICLR}, 2018.

\bibitem{he_deep_2016}
Kaiming He, Xiangyu Zhang, Shaoqing Ren, and Jian Sun.
\newblock {Deep Residual Learning for Image Recognition}.
\newblock In {\em Proc. of the CVPR}, 2016.

\bibitem{huster_limitations_2018}
Todd Huster, Cho-Yu~Jason Chiang, and Ritu Chadha.
\newblock {{Limitations of the {Lipschitz} Constant as a Defense against
  Adversarial Examples}}.
\newblock In {\em ECML PKDD}, pages 16--29, 2018.

\bibitem{kannan_adversarial_2018}
Harini Kannan, Alexey Kurakin, and Ian~J. Goodfellow.
\newblock {{Adversarial {Logit} {Pairing}}}.
\newblock {\em CoRR}, abs/1803.06373, 2018.

\bibitem{kopuklu2019convolutional}
Okan K{\"o}p{\"u}kl{\"u}, Maryam Babaee, Stefan H{\"o}rmann, and Gerhard
  Rigoll.
\newblock {{Convolutional Neural Networks with Layer Reuse}}.
\newblock In {\em Proc. of ICIP}, 2019.

\bibitem{kurakin_adversarial_2016}
Alexey Kurakin, Ian~J. Goodfellow, and Samy Bengio.
\newblock {{Adversarial Examples in the Physical World}}.
\newblock {\em CoRR}, abs/1607.02533, 2016.

\bibitem{lin_feature_2017}
Tsung-Yi Lin, Piotr Doll{\'a}r, Ross~B. Girshick, Kaiming He, Bharath
  Hariharan, and Serge~J. Belongie.
\newblock {Feature {Pyramid} {Networks} for {Object} {Detection}}.
\newblock In {\em Proc. of the CVPR}, pages 936--944, 2017.

\bibitem{madry_towards_2017}
Aleksander Madry, Aleksandar Makelov, Ludwig Schmidt, Dimitris Tsipras, and
  Adrian Vladu.
\newblock {{Towards {Deep} {Learning} {Models} {Resistant} to {Adversarial}
  {Attacks}}}.
\newblock {\em CoRR}, abs/1706.06083, 2017.

\bibitem{meng_magnet:_2017}
Dongyu Meng and Hao Chen.
\newblock {{Magnet: a Two-pronged Defense against Adversarial Examples}}.
\newblock In {\em Proc. of the SIGSAC}, pages 135--147, 2017.

\bibitem{moosavi-dezfooli_deepfool:_2016}
Seyed-Mohsen Moosavi-Dezfooli, Alhussein Fawzi, and Pascal Frossard.
\newblock {{Deepfool: a Simple and Accurate Method to Fool Deep Neural
  Networks}}.
\newblock In {\em {Proc. of the CVPR}}, pages 2574--2582, 2016.

\bibitem{nguyen_deep_2015}
Anh Nguyen, Jason Yosinski, and Jeff Clune.
\newblock {{Deep Neural Networks are Easily Fooled: {High} Confidence
  Predictions for Unrecognizable Images}}.
\newblock In {\em {Proc. of the CVPR}}, pages 427--436, 2015.

\bibitem{nicolae_adversarial_2018}
Maria-Irina Nicolae, Mathieu Sinn, Minh~Ngoc Tran, Beat Buesser, Ambrish Rawat,
  Martin Wistuba, Valentina Zantedeschi, Nathalie Baracaldo, Bryant Chen, Heiko
  Ludwig, Ian Molloy, and Ben Edwards.
\newblock {{Adversarial {Robustness} {Toolbox} v0.10.0}}.
\newblock {\em CoRR}, abs/1807.01069, 2018.

\bibitem{papernot_practical_2017}
Nicolas Papernot, Patrick~D. McDaniel, Ian~J. Goodfellow, Somesh Jha, Z.~Berkay
  Celik, and Ananthram Swami.
\newblock {{Practical {Black}-{Box} {Attacks} against {Machine} {Learning}}}.
\newblock In {\em Proc. of the AsiaCCS}, pages 506--519, 2017.

\bibitem{rudin_nonlinear_1992}
Leonid~I Rudin, Stanley Osher, and Emad Fatemi.
\newblock {{Nonlinear Total Variation Based Noise Removal Algorithms}}.
\newblock {\em Physica D: nonlinear phenomena}, 60(1-4):259--268, 1992.

\bibitem{song_pixeldefend:_2018}
Yang Song, Taesup Kim, Sebastian Nowozin, Stefano Ermon, and Nate Kushman.
\newblock {{{PixelDefend}: {Leveraging} {Generative} {Models} to {Understand}
  and {Defend} against {Adversarial} {Examples}}}.
\newblock In {\em {Proc. of the ICLR}}, 2018.

\bibitem{szegedy_intriguing_2013}
Christian Szegedy, Wojciech Zaremba, Ilya Sutskever, Joan Bruna, Dumitru Erhan,
  Ian Goodfellow, and Rob Fergus.
\newblock {{Intriguing Properties of Neural Networks}}.
\newblock {\em CoRR}, abs/1312.6199, 2013.

\bibitem{tramer_ensemble_2018}
Florian Tram{\`e}r, Alexey Kurakin, Nicolas Papernot, Ian Goodfellow, Dan
  Boneh, and Patrick McDaniel.
\newblock {{Ensemble {Adversarial} {Training}: {Attacks} and {Defenses}}}.
\newblock In {\em {Proc. of the ICLR}}, 2018.

\bibitem{vaswani_attention_2017}
Ashish Vaswani, Noam Shazeer, Niki Parmar, Jakob Uszkoreit, Llion Jones,
  Aidan~N. Gomez, Lukasz Kaiser, and Illia Polosukhin.
\newblock {Attention is All you Need}.
\newblock In {\em Proc. of the NIPS}, 2017.

\bibitem{xie_feature_2019}
Cihang Xie, Yuxin Wu, Laurens van~der Maaten, Alan~L Yuille, and Kaiming He.
\newblock {{Feature Denoising for Improving Adversarial Robustness}}.
\newblock In {\em {Proc. of the CVPR}}, pages 501--509, 2019.

\bibitem{xie_aggregated_2017}
Saining Xie, Ross~B. Girshick, Piotr Doll{\'a}r, Zhuowen Tu, and Kaiming He.
\newblock {{Aggregated {Residual} {Transformations} for {Deep} {Neural}
  {Networks}}}.
\newblock In {\em {Proc. of the CVPR}}, pages 5987--5995, 2017.

\bibitem{xu_interpreting_2019}
Kaidi Xu, Sijia Liu, Gaoyuan Zhang, Mengshu Sun, Pu Zhao, Quanfu Fan, Chuang
  Gan, and Xue Lin.
\newblock {{Interpreting {Adversarial} {Examples} by {Activation} {Promotion}
  and {Suppression}}}.
\newblock {\em CoRR}, abs/1904.02057, 2019.

\bibitem{yan_deep_2018}
Ziang Yan, Yiwen Guo, and Changshui Zhang.
\newblock {{Deep {Defense}: {Training} {DNNs} with {Improved} {Adversarial}
  {Robustness}}}.
\newblock In {\em {Proc. of the NIPS}}, pages 419--428, 2018.

\end{thebibliography}
}

\end{document}